\documentclass{article} 

\usepackage[x11names]{xcolor}
\usepackage{hyperref}
\usepackage{enumitem}
\usepackage{url}
\usepackage{lipsum,lmodern}
\usepackage{tikz}
\usepackage{subcaption}
\usepackage{amsthm} 
\usepackage{footnote}
\usepackage{booktabs} 
\usepackage{threeparttable}  
\usepackage{multirow} 
\usepackage{fancyhdr}
\usetikzlibrary{arrows,calc,decorations.pathmorphing,arrows.meta}

\usepackage[most]{tcolorbox}
\usepackage{array}      
\usepackage{iclr2025_conference,times}
\iclrfinalcopy

\usepackage{amsmath,amsfonts,bm}

\def\eqref#1{equation~\ref{#1}}

\def\1{\bm{1}}

\DeclareMathAlphabet{\mathsfit}{\encodingdefault}{\sfdefault}{m}{sl}
\SetMathAlphabet{\mathsfit}{bold}{\encodingdefault}{\sfdefault}{bx}{n}

\hypersetup{
    colorlinks=true,
    linkcolor=blue,         
    citecolor=green,       
    urlcolor=cyan
}

\newcommand*\circled[1]{\tikz[baseline=(char.base)]{
            \node[shape=circle,draw,inner sep=0.5pt] (char) {#1};}}

\theoremstyle{plain}     

\newtheorem{proposition}{Proposition}

\theoremstyle{remark}

\usepackage[nameinlink,capitalise]{cleveref}

\crefname{section}{Sec.}{Sec.}
\crefname{line}{line}{§§}
\crefname{figure}{Fig.}{Fig.}
\crefname{table}{Tab.}{§§}
\crefname{algorithm}{Algorithm}{§§}
\crefname{appendix}{Appx.}{§§}
\crefname{definition}{Def.}{§§}
\crefname{equation}{Eq.}{Eq.}
\crefname{step}{Step.}{§§}
\crefname{remark}{Remark}{§§}
\crefname{paragraph}{Para.}{§§}
\crefname{theorem}{Thm.}{§§}
\crefname{proposition}{Propos.}{§§}

\newcommand{\coloredcref}[1]{\textcolor{blue}{\cref{#1}}}

\title{How can representation dimension dominate structurally pruned LLMs?}

\author{Mingxue Xu, Lisa Alazraki \& Danilo P. Mandic \\
Imperial College London\\
London SW7 2AZ, United Kingdom  \\
\texttt{\{m.xu21, lisa.alazraki20, d.mandic\}@imperial.ac.uk} \\
}

\begin{document}

\maketitle
\begin{abstract}
Pruning assumes a subnetwork exists in the original deep neural network~\citep{franklelottery}, which can achieve comparative model performance with less computation than the original. 
However, it is unclear how the model performance varies with the different subnetwork extractions.
In this paper, we choose the representation dimension (or embedding dimension, model dimension, the dimension of the residual stream in the relevant literature) as the entry point to this issue.
We investigate the linear transformations in the LLM transformer blocks and consider a specific structured pruning approach, SliceGPT~\citep{slicegpt}, to extract the subnetworks of different representation dimensions.
We mechanistically analyse the activation flow during the model forward passes, and find the representation dimension dominates the linear transformations, model predictions, and, finally, the model performance.
Explicit analytical relations are given to calculate the pruned model performance (perplexity and accuracy) without actual evaluation, and are empirically validated with Llama-3-8B-Instruct and Phi-3-mini-4k-Instruct.

\end{abstract}

\section{Introduction}
Recent progress in pruning for LLMs has proved that freezing or deleting unnecessary LLM model weights can retain similar language task performances with less computations~\citep{sparsegpt,men2025shortgpt}. However, current research has not fully addressed the functionality shifts after pruning, which may cause safety issues (e.g. model collapse~\citep{yang2024fall} and unknown backdoor features~\citep{wang2024transtroj}), and makes it difficult to set pruning hyperparameters (e.g., sparsity). 

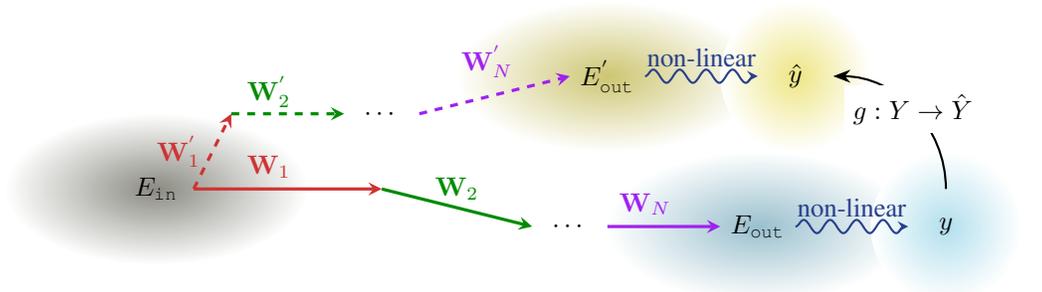
\begin{figure}[h]
    \centering
\begin{tikzpicture}
\coordinate (O1) at (0,0); 
\coordinate (O2) at (8,-0.5); 
\coordinate (O3) at (6,1.5); 
\coordinate (R2) at (8.5,1.5);
\coordinate (R1) at (10.5,-0.5);

\shade[outer color=white, inner color=Ivory4] (O1) ellipse (2cm and 1cm);
\shade[outer color=white, inner color=LightBlue3] (O2) ellipse (2cm and 1cm);
\shade[outer color=white, inner color=Khaki3] (O3) ellipse (2cm and 1cm);
\shade[inner color=LightBlue2] (R1) ellipse (1cm and 1cm);
\shade[ inner color=Khaki2] (R2) ellipse (1cm and 1cm);

\node at (O1) {$E_{\texttt{in}}$};
\node at (O2) {$E_{\texttt{out}}$};
\node at (O3) {${E}^{'}_{\texttt{out}}$};
\node at (R2) {$\hat{y}$};
\node at (R1) {$y$};

\node[color=Brown3] at ($(O1.center)+(1.5,0.3)$) {${\bf W}_1$};
\node[color=Green4] at ($(O1.center)+(4,0)$) {${\bf W}_2$};
\node[color=Purple0] at ($(O1.center)+(6.5,-0.2)$) {${\bf W}_{N}$};

\node[color=Brown3] at ($(O1.center)+(0.3,0.5)$) {${\bf W}^{'}_1$};
\node[color=Green4] at ($(O1.center)+(1.5,1.3)$) {${\bf W}^{'}_2$};
\node[color=Purple0] at ($(O1.center)+(4.4,1.7)$) {${\bf W}^{'}_{N}$};


\draw[-stealth,decorate,decoration=snake, thick,color=RoyalBlue4] ($(O2.center)+(0.5,0)$) -- node[above] {non-linear} ($(R1.center)+(-0.5,0)$);
\draw[-stealth,decorate,decoration=snake, thick,color=RoyalBlue4] ($(O3.center)+(0.5,0)$) -- node[above] {non-linear} ($(R2.center)+(-0.5,0)$);

\draw[-stealth, very thick, color=Brown3] ($(O1.center)+(0.5,0)$) -- ($(O1.center)+(3,0)$);
\draw[-stealth, very thick, dashed, color=Brown3] ($(O1.center)+(0.5,0)$) -- ($(O1.center)+(1,1)$);

\draw[-stealth, very thick, color=Green4] ($(O1.center)+(3,0)$) -- ($(O1.center)+(5,-0.5)$);
\draw[-stealth, very thick, dashed, color=Green4] ($(O1.center)+(1,1)$) -- ($(O1.center)+(2.5,1)$);

\node at ($(O1.center)+(3,1)$) {$\cdots$}; 
\node at ($(O1.center)+(5.5,-0.5)$) {$\cdots$}; 

\draw[-stealth, very thick, color=Purple0] ($(O1.center)+(6,-0.5)$) -- ($(O2.center)+(-0.5,0)$);
\draw[-stealth, very thick, dashed, color=Purple0] ($(O1.center)+(3.5,1)$) -- ($(O3.center)+(-0.5,0)$);


\draw[thick,-{Stealth[scale=1]}] ($(R1.center)+(0,0.5)$)  to[bend right=45]
        node[fill=white] {$g: Y \rightarrow \hat{Y}$} ($(R2.center)+(0.5,0)$);

\end{tikzpicture}
\vspace{-5pt}
    \caption{The linear transformations before (solid arrows) and after (dashed arrows) pruning in the model. 
    The original model maps the input embeddings $E_{\texttt{in}}$ from the input space (illustrated with the shading grey ellipse) to the output embedding $E_{\texttt{out}}$ in the output space, through a series of linear transformations (i.e. those defined by the weight matrices ${\bf W}_1, \ldots,{\bf W}_N$ ). After pruning, ${\bf W}_i$ is converted to ${\bf W}^{'}_i$ ($i \in \{1,2,\ldots, N\}$), and the original output $E_{\texttt{out}}$ is shifted to $E^{'}_{\texttt{out}}$. 
    The final predictions ($y$ and $\hat{y}$) are generated (normally non-linearly) from $E_{\texttt{out}}$ and $E^{'}_{\texttt{out}}$. Denote the prediction domain of the pruned model as $\hat{Y}$ and that of the unpruned model $Y$, the mapping $g:Y\rightarrow\hat{Y}$ shifts the model performance.}
    \label{fig:idea}
\end{figure}

To investigate this functionality shifting issue, we take the LLM as a self-contained system. We analyse the mappings from the input space to the output space, how pruning transforms these mappings, and then transforms the model predictions, as shown in~\cref{fig:idea}. This analysis compiles well with structured pruning on LLM transformers since \circled{1} structured pruning works on structural components, which means it edits the linear representations and linear transformations in LLMs~\citep{park2023linear}, then rewrites the transmitted signals throughout the network; \circled{2} the extracted subnetworks are also the subgraphs in the model computational graph with distinct functionality, so they are human-understandable ``circuits''~\citep{wanginterpretability}. 

Focusing on the representation dimension in LLMs, our contributions are as follows:
\begin{enumerate}[leftmargin=*]
    \item We carry out a mechanistic investigation and find that the representation dimension dominates the transformations (linear and non-linear) in the LLM forward passes. This dominance includes global input-output mappings and how the weight matrices of different dimensions interact with the activation flow locally.
    \item With the definition of the model performance metrics (i.e. perplexity and multiple-choice accuracy), we give the analytical relations to quantify the pruned model performance with sparsity. We verify our analytical relations empirically with two LLMs, LlaMa-3 and Phi-3, using SliceGPT~\citep{slicegpt} for dimension reduction.
    
\end{enumerate}
\vspace{-5pt}

\vspace{-10pt}
\section{Activation flow in LLM Transformers}\label{sec:act_flow}

\begin{figure}
    \centering
    \begin{tikzpicture}[scale=0.85]

\coordinate (E_in) at (0.5,1.5); 
\coordinate (W_norm) at ($(E_in.center)+(2,-0.25)$);
\coordinate (W_k) at ($(E_in.center)+(4,-0.25)$); 
\coordinate (softmax) at ($(W_norm)+(3.5,0.5)$); 
\coordinate (mlp_start) at ($(W_norm)+(5,0)$); 

\fill[fill=Azure3, fill opacity=0.4] (1.9,0.4) rectangle (13.5,3);
\draw[-stealth] ($(mlp_start.center)+(5,-0.4)$) -- ($(mlp_start.center)+(6,0)$);
\draw[-stealth] ($(E_in.center)+(0.75,0)$) -- ($(E_in.center)+(1.9,0.2)$);

\fill[fill=Azure3, fill opacity=0.4] (1.7,0.2) rectangle (13.3,2.8);

\draw[-stealth] ($(mlp_start.center)+(5,-0.6)$) -- ($(mlp_start.center)+(6,0)$);
\draw[-stealth] ($(E_in.center)+(0.75,0)$) -- ($(E_in.center)+(1.7,0)$);

\fill[fill=Azure3, fill opacity=0.9] (1.5,0) rectangle (13.1,2.6);
\draw[-stealth] ($(E_in.center)+(0.75,0)$) -- ($(E_in.center)+(1.4,-0.2)$);

\node at ($(E_in.center)+(0.5,0)$) {$E_{\texttt{in}}$};
\node at (W_norm) {${\bf W}^{(1)}_{\texttt{norm}}$};

\node[color=DeepSkyBlue4] at ($(E_in.center)+(2,-1.25)$) {$\texttt{RMSNorm}(\cdot)$};

\node at ($(E_in.center)+(4,0.75)$) {${\bf W}_{q}$};

\node at (W_k) {${\bf W}_{k}$};

\node at ($(E_in.center)+(4,-1.25)$) {${\bf W}_{v}$};

\node[color=DeepSkyBlue4] at (softmax) {$\texttt{softmax}(\cdot)$};


\draw[-stealth] ($(W_norm.center)+(0,-0.25)$) -- ($(W_norm.center)+(0,-0.75)$);
\draw[-stealth] ($(W_norm.center)+(1,-1)$) -- ($(W_norm.center)+(1.6,1)$);
\draw[-stealth] ($(W_norm.center)+(1,-1)$) -- ($(W_norm.center)+(1.6,0)$);
\draw[-stealth] ($(W_norm.center)+(1,-1)$) -- ($(W_norm.center)+(1.6,-1)$);

\draw[-stealth] ($(softmax.center)+(-1.1,-0.5)$) -- ($(softmax.center)+(-0.2,-0.25)$);
\draw[-stealth] ($(softmax.center)+(-1.1,0.5)$) -- ($(softmax.center)+(-0.2,0.25)$);

\draw[-stealth] ($(softmax.center)+(2.9,-1.5)$) -- ($(mlp_start)+(2.5,-1)$);
\draw[-stealth] ($(softmax.center)+(2.9,-1.5)$) to[bend left=30] ($(mlp_start)+(2.5,1)$);
\draw[-stealth] ($(softmax.center)+(-1.1,-1.5)$) to[bend right=45] ($(softmax.center)+(1.1,0)$);
\draw[-stealth] ($(softmax.center)+(1.1,0)$) -- ($(softmax.center)+(1.6,0)$);

\node at ($(softmax.center)+(2.2,0)$) {${\bf W}^{(2)}_{\texttt{norm}}$};
\draw[-stealth] ($(softmax.center)+(2.2,-0.3)$) -- ($(softmax.center)+(2.2,-1.2)$);
\node[color=DeepSkyBlue4] at ($(softmax.center)+(1.9,-1.5)$) {$\texttt{RMSNorm}(\cdot)$};
\node at ($(mlp_start)+(3,1)$) {${\bf W}_\texttt{gate}$};
\node at ($(mlp_start)+(3,-1)$) {${\bf W}_\texttt{up}$};
\node at ($(mlp_start)+(5,-1)$) {${\bf W}_\texttt{down}$};

\node[color=DeepSkyBlue4] at ($(mlp_start.center)+(3,0)$) {$\texttt{Hadamard}(\cdot)$};

\draw[-stealth] ($(mlp_start)+(2.8,0.75)$) -- ($(mlp_start)+(2.8,0.25)$);
\draw[-stealth] ($(mlp_start)+(2.8,-0.75)$) -- ($(mlp_start)+(2.8,-0.25)$);
\draw[-stealth] ($(mlp_start)+(4.8,-0.25)$) -- ($(mlp_start)+(4.8,-0.75)$);

\draw[-stealth] ($(mlp_start.center)+(4,0)$) -- ($(mlp_start.center)+(4.4,0)$);
\node[color=DeepSkyBlue4] at ($(mlp_start.center)+(5,0)$) {$\sigma(\cdot)$};

\node at ($(mlp_start.center)+(6.2,0)$) {$\bigoplus$};

\node at ($(mlp_start.center)+(7.5,0)$) {${\bf W}_{\texttt{norm}}$};
\node[color=DeepSkyBlue4] at ($(softmax.center)+(9,-1.5)$) {$\texttt{RMSNorm}(\cdot)$};

\draw[-stealth] ($(mlp_start)+(7.5,-0.25)$) -- ($(mlp_start)+(7.5,-0.75)$);

\draw[-stealth] ($(mlp_start.center)+(5,-0.8)$) -- ($(mlp_start.center)+(6,0)$);

\draw[-stealth] ($(mlp_start.center)+(6.5,0)$) -- ($(mlp_start.center)+(7,0)$);

\draw[-stealth] ($(mlp_start)+(7.8,-0.75)$) -- ($(mlp_start.center)+(8.5,0)$);

\node at ($(mlp_start.center)+(9,0)$) {$E_{\texttt{out}}$};

\end{tikzpicture}
    \caption{Activation flow in the LLM transformer. Linear transformations are defined by weight matrices like ${\bf W}_i$, and non-linear transformations are represented with \texttt{\textcolor{DeepSkyBlue4}{teletype font}}. The detailed shapes of the weight matrices are clarified in~\cref{tab:shape}.}
    \label{fig:act_flow}
\end{figure}
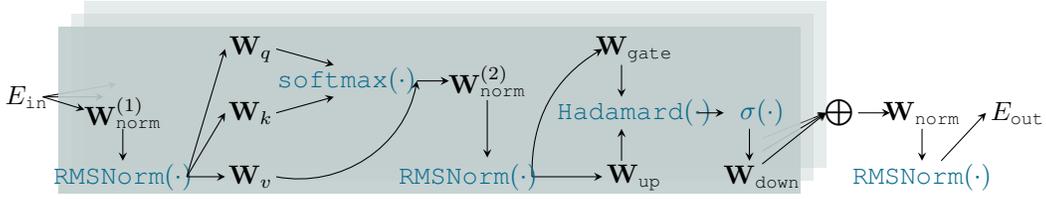

This section clarifies how the linear representations are transmitted throughout the network, so that we can analyse the impacts of dimension reduction on this transmission in~\cref{sec:pruning}. 
We use the transformer architecture structure in~\cite{attn}, denoting the LLM transformer module as $\mathcal{M}$. $\mathcal{M}$ contains groups of linear and non-linear transformations in multi-head attention implementation, as shown in~\cref{fig:act_flow}.

\begin{proposition} \label{pro:transformation}
    The mapping $ M: { E}_{\texttt{in}} \rightarrow \bigoplus_{i=1}^h A_i$ defined by the transformer $\mathcal{M}$, consists of $h$ groups of the following transformations:
    \vspace{-5pt}
    \begin{itemize}[leftmargin=*]
        \item {\bf linear transformations} defined by $\{{\bf W}_{\texttt{norm}},{\bf W}_{q},{\bf W}_{k},{\bf W}_{v},{\bf W}_{o},{\bf W}_{\texttt{gate}},{\bf W}_{\texttt{up}},{\bf W}_{\texttt{down}}\}$;
        \vspace{-5pt}
        \item {\bf non-linear transformations} involved in $\{\texttt{RMSNorm}(\cdot),\texttt{softmax}(\cdot), \sigma(\cdot)\}$.
    \end{itemize}
\end{proposition}


With the transformations in~\cref{pro:transformation}, we have the following analysis of the activation flow.  

Let $d$ denote the representation dimension. We input a text of $l$ tokens to the embedding layer and obtain the encoded texts (embedding matrix) ${ E}_{\texttt{in}} \in \mathbb{R}^{l \times d}$. $ E_{\texttt{in}}$ is then processed by a series of linear and non-linear transformations, as clarified in~\cref{sec:act}.

Suppose there are $h$ parallel attention layers in $\mathcal{M}$, the attention outputs of the $i$th attention layer is $A_i$, so the final output of the transformer module $\mathcal{M}$ is 
\begin{equation}
    \bigoplus_{i=1}^h A_i = A_1 \oplus A_2 \oplus \cdots \oplus A_h \in \mathbb{R}^{h \times l \times h_{\texttt{attn}}h_{\texttt{dim}}}.
\end{equation}
Here, $\oplus$ denotes the direct sum, which is equivalent to concatenation in the actual implementation~\citep{xu2024geometry,barbero2024round}. Since in the default setting, $h_{\texttt{attn}}h_{\texttt{dim}}=d$, we have the final hidden states output $\bigoplus_{i=1}^h A_i \in \mathbb{R}^{h \times l \times d} $.

We can observe that most of this transmission implemented by the transformer is linear transformations, which are also the investigated objects in the scaling law~\citep{kaplan2020scaling}. {\bf The representation dimension $d$ dominates the linear transformation throughout the whole network}.

\vspace{-10pt}
\section{Functionality Shifting caused by Dimension Reduction}\label{sec:pruning}

In this section, we discuss the functionality shifting caused by the structured pruning, i.e., the mapping from the unpruned model prediction domain to that of the pruned model ($g: Y \rightarrow \hat{Y}$ in~\cref{fig:idea}), and then the shift in the model performance. 
We adopt SliceGPT as the structure pruning approach, since this approach unifiedly reduces the representation dimension in the attention and feedforward layers. We denote sparsity as $s$ ($0<s<1$), $(1-s)d\in \mathbb{Z}^{+}$ the representation dimension after pruning.

\subsection{Perplexity}

Perplexity, denoted as $\texttt{PPL}(\cdot)$, represents the uncertainty of a discrete probability distribution. Let us start with a single token sequence $X = (x_1,x_2,\ldots,x_l)$. We assign each token in $X$ to an embedding vector ${\bf v}_i = (v_{i,1}, v_{i,2}, \ldots, v_{i,d}) \in \mathbb{R}^{1 \times d}$, and obtain the stacked embedding vectors (embedding matrix) $E=({\bf v}_1,{\bf v}_2,\ldots, {\bf v}_l)\in \mathbb{R}^{l  \times d}$. The differential entropy of the stacked embedding vectors is $H(E_\texttt{in}) = H(E_\texttt{out}) = ld \cdot \kappa(t)$, where $\kappa(t) = -\int p(t)\log(p(t))\;\; dt$. After pruning, the entropy of the output embeddings shifts from $ld \cdot \kappa(t)$ to $(1-s)ld \cdot \kappa(t)$, such that $\frac{H({E^{'}_{\texttt{out}}})}{H({E_{\texttt{out}}})}={1-s}$. Given the definition of perplexity correlation is $\texttt{PPL}(E)= 2^{H(E)}$, we have the following theorem about the perplexity before and after pruning:

\begin{proposition}\label{th:ppl} For structured pruning with representation dimension sparsity $s$, the perplexity of the pruned model, denoted as $\texttt{PPL}(D)$, and the perplexity of the original model, denoted as $\texttt{PPL}_0(D)$, satisfy
\begin{equation}\label{eq:ppl}
  \frac{ \ln \texttt{PPL}_0(D)}{ \ln \texttt{PPL}(D)}  = 1-s , 
\end{equation}
where $D$ is the test dataset from the same distribution as the dataset used for pruning.
\end{proposition}

To verify~\cref{th:ppl}, we pruned Llama-3-8B-Instruct~\citep{llama} and Phi-3-mini-4k-Instruct~\citep{phi} with different dimension sparsity with SliceGPT, then evaluated the pruned and unpruned models on WikiText2, as shown in~\cref{fig:ppl}. We can observe that the perplexities fit well with the linear function, which justifies our analytical expression~\cref{eq:ppl}. 

\subsection{Accuracy}


Accuracy of generative language models is bonded to the sequence probability, while perplexity is defined as the multiplicative inverse of the sequence probability. A well-generalized model would assign a high probability to the correct sequence (high accuracy but low perplexity). Therefore, intuitively, accuracy is inversely relevant to perplexity.
 
\begin{figure}[t]
     \centering
     \begin{subfigure}[b]{0.32\textwidth}
         \centering
         \includegraphics[width=\textwidth]{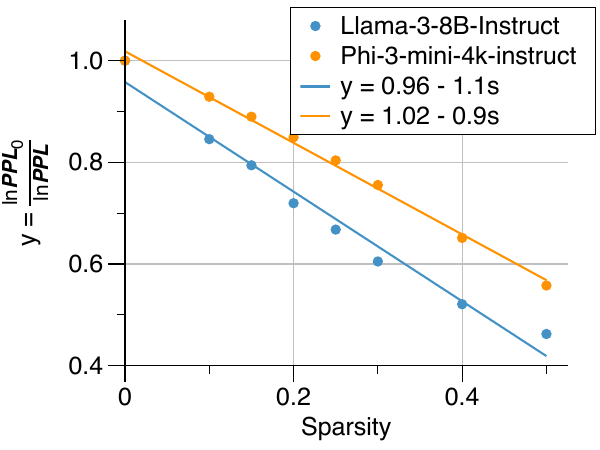}
         \caption{Perplexity on WikiText2}
         \label{fig:ppl}
     \end{subfigure}
          \hfill
     \begin{subfigure}[b]{0.67\textwidth}
         \centering
         \includegraphics[width=\textwidth]{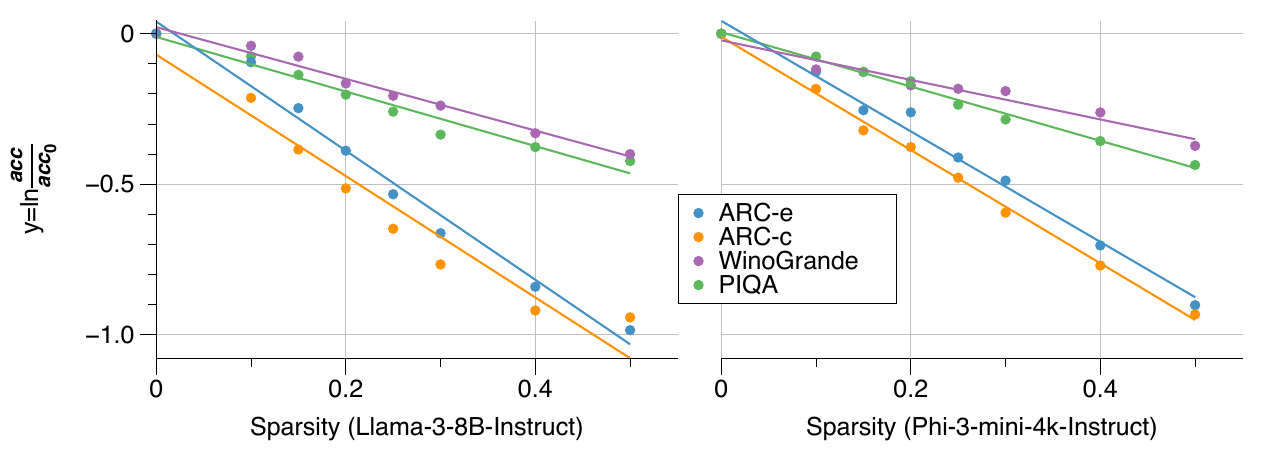}
         \caption{Accuracy in binary/multiple-choice Tasks}
         \label{fig:acc}
     \end{subfigure}
    \caption{Fitted evaluation results. The fitting coefficients and errors are in~\cref{sec:error}.}
    \vspace{-10pt}
    \label{fig:eval}
\end{figure}

We postulate that a logarithmic form like~\cref{eq:ppl} might also exist in the correlation between accuracy and representation dimension sparsity. Note that here we consider the accuracy of short, multiple-choice answers (e.g. “Yes”/“No”, or “A”/“B”/“C”), since this accuracy can be computed via exact match without ambiguity. We tried possible logarithmic expressions similar to $\frac{\ln \texttt{PPL}_0 (D)}{\ln \texttt{PPL}(D)}$ with the same LLMs used for perplexity evaluation. We replaced the perplexities with multiple-choice accuracies on ARC-e, ARC-c, WinoGrande and PIQA. Among these logarithmic expressions, the best-fitted form is $\ln \frac{ {\texttt{acc}}}{\texttt{acc}_0}$, as shown in~\cref{fig:acc}. Thus we have the following empirical theorem with similar analytical expressions as~\cref{th:ppl}:

\begin{proposition}
For structured pruning with representation dimension sparsity $s$, the multiple-choice accuracy of the pruned model, denoted as $\texttt{acc}(D)$, and that of the original model, denoted as $\texttt{acc}_0(D)$, satisfy
\begin{equation}\label{eq:acc}
  \ln \frac{\texttt{acc}(D)}{ \texttt{acc}_0(D)}  \propto 1-s , 
\end{equation}
where $D$ is the test dataset for evaluating multiple-choice accuracy.

\end{proposition}\label{th:acc}

We can observe that the left-hand side of~\cref{eq:acc} is almost the multiplicative inverse of the left-hand side of~\cref{eq:ppl}. This is consistent with the fact that $ \frac{1}{\texttt{PPL}}$ and ${\texttt{acc}}$ are positively correlated when the model generalizes well, and they have (nearly) the same range ($ \frac{1}{\texttt{PPL}} \in (0,1]$ and $ {\texttt{acc}} \in [0,1]$). 
The right-hand sides of~\cref{eq:ppl,eq:acc} are exactly the same, indicating that the representation dimension dominates the model predictions. Though the multiple-choice accuracies fit~\cref{eq:acc} albeit less accurately than the perplexities, this formula consistency further empirically justifies what we claimed in~\cref{sec:act_flow}.



\section{Related Work}
\paragraph{Sparsity and Scaling Law in LLM.}
Sparsifying LLMs at the matrix level~\citep{sparsegpt}, module level~\citep{men2025shortgpt} or representation dimension level~\citep{slicegpt}, has drawn a lot of attention in recent years.
The correlation between sparsity and model performance has remained underexplored. In this work, we analytically investigate this correlation.
~\cite{frantar2024scaling} put sparsity into the original scaling law~\citep{kaplan2020scaling}, but consider other factors, such as the size of the dataset and training steps. In the context of scaling law, we solely focus on the entropy loss shifting with sparsity, and our~\cref{eq:ppl,eq:acc} are expressed in standard evaluation metrics. Broadly speaking, our work is a special case of scaling law, but finer-grained, more precise and more analytical than the original scaling law.  

\paragraph{Circuits in Transformers.} In the field of mechanistic interpretability, circuits are the subgraphs in the neural networks that have distinct computation mechanisms~\citep{jainmechanistically,Arthur}. In the context of the circuit analysis in transformers,~\cite{elhage2021mathematical} discusses the contributions of the weight matrices to the final residual stream from a mathematical perspective. However, they do not fully address the impacts of the representation dimension on the model predictions. We dive into these impacts and give quantitative conclusions on two common evaluation metrics (i.e. perplexity and accuracy).  
\section{Conclusion}
This work mechanistically investigates the impacts of representation dimension on the model predictions and, herein, model performance. It introduces analytical relations to estimate the pruned model performance, which is empirically valid in real-world LLM settings. Limitations and future work are discussed in~\cref{app:future}.

\bibliographystyle{iclr2025_conference}
\bibliography{ref}

\newpage
\appendix
\section{Appendix}

\begin{table}[h]
    \centering
        \caption{Notation in this paper.}\label{tab:notation}
    \begin{tabular}{l|l}
    \toprule
    {\bf Symbol} & {\bf Meaning}  \\
    \midrule
    $\oplus$ & Direct sum (concatenation in practice). \\
    $d$ &       Embedding dimension, or model dimension~\citep{kaplan2020scaling},\\ & the dimension of the  residual stream \\ & ~\citep{elhage2021mathematical,kaplan2020scaling}\\
    $m$ & Intermediate size of the MLPs.\\
    $h_{\texttt{dim}}$ &   Dimension of the attention heads.   \\
    $h_{\texttt{attn}}$ &  Number of the attention heads.     \\
    $v$ &  Number of the key-value heads.    \\
    $h$ & Number of the attention layers.     \\
    $l$ & Token sequence length.     \\
    $s$ &   Embedding dimension sparsity.   \\
    $\gamma$ & Scaling factor.     \\
    ${\bf v}$, ${\bf v}_i$ & Embedding vector, $i$th row the embedding matrix $E$.     \\
    ${\bf W}$ &  Weight matrix.    \\
    ${\mathcal{M}}$ &  Transformer module.    \\
    $D$ & Dataset. \\
    $N$ & The number of the layers to be compressed in $\mathcal{M}$. \\
    $E$, $E_{\texttt{in}}$, $E_{\texttt{out}}$  & Embedding matrix, input embedding matrix, output \\ & embedding matrix. \\
    $A$, $A_i$  & Output of the attention layer, output of the $i$th attention layer.  \\
    $H$ & Entropy. \\
    $\kappa$ & Unit entropy of a real scalar variable. \\
    $\sigma$ & Activation function. \\  
    
         \bottomrule
    \end{tabular}
    
\end{table}

\begin{table}[h]
\centering
\begin{threeparttable}
\caption{The weight matrices in the transformer modules and their sizes, ordered by their appearances during the forwarding pass. The detailed notation is in~\coloredcref{tab:notation}.}
\label{tab:shape}
\begin{tabular}{l|l|l|l||l|l}

\toprule
${\bf W}_{\texttt{RMSNorm}}$ & $d$                               & ${\bf W}_{v}$ & $d \times v h_{\texttt{dim}}$    &           ${\bf W}_{\texttt{up}}$ & $d \times m$ \\ \hline
 ${\bf W}_{q}$ & $d \times h_{\texttt{attn}}h_{\texttt{dim}}$ \tnote{$\star$}   & ${\bf W}_{o}$ & $h_{\texttt{attn}}h_{\texttt{dim}} \times d$                & ${\bf W}_{\texttt{down}}$ & $m \times d$ \\ \hline
${\bf W}_{k}$ & $d \times v  h_{\texttt{dim}}$                                                 & ${\bf W}_{\texttt{gate}}$  & $d \times m$ & \\
\bottomrule 

\end{tabular}
\begin{tablenotes}
            \small
            \item[$\star$] $h_{\texttt{attn}}h_{\texttt{dim}}=d$
        \end{tablenotes}
\end{threeparttable}
\end{table}


    
\subsection{Activation Flow alysis}\label{sec:act}

Let $d$ be the embedding dimension. Input a text of $l$ tokens to the embedding layer first, we then have the encoded texts (embeddings) ${ E}_{\texttt{in}} \in \mathbb{R}^{l \times d}$. $ E_{\texttt{in}}$ is then processed by a layer norm operation. Take the current widely-used RMSNorm~\citep{zhang2019root} as our case, the normalized embeddings after the layer norm operation is 
\begin{align}
\bar{ E}_{\texttt{in}}  &= {\bf W}_{\texttt{norm}} { E}_{\texttt{in}} \cdot \texttt{RMSNorm}({ E}_{\texttt{in}})\in \mathbb{R}^{l \times d},\\
\texttt{RMSNorm}({ E}) &= {\sqrt{\frac{1}{d}\sum_{i=1}^d { E}^{\top}[:,i]{ E}[:,i]}} \in \mathbb{R} ,
\end{align}
where ${\bf W}_{\texttt{RMSNorm}}\in \mathbb{R}^{d}$ is the weight of the RMSNorm layer, ${ E}_{\texttt{in}}[:, i]$ is the $i$th column of ${ E}_{\texttt{in}}$.

 The attention layer typically consists of four linear layers with weight matrices ${\bf W}_q\in \mathbb{R}^{d \times h_{\texttt{attn}}h_{\texttt{dim}}}$,  ${\bf W}_k\in \mathbb{R}^{d \times vh_{\texttt{dim}}}$,  ${\bf W}_v\in \mathbb{R}^{d \times vh_{\texttt{dim}}}$ and ${\bf W}_o\in \mathbb{R}^{h_{\texttt{attn}}h_{\texttt{dim}} \times d}$. $h_{\texttt{attn}}$ is the number of the attention heads, $h_{\texttt{dim}}$ is the head dimension, and $v$ is the number of key-value heads\footnote{We use the same notation names as listed in \href{https://huggingface.co/meta-llama/Meta-Llama-3-8B/blob/main/config.json}{Llama-3-8B Configuration}.}. 
 In the default setting of Llama\footnote{For simplicity, we do not consider position encoding, dropout, attention mask, and MLP bias in LLMs.}, $h_{\texttt{attn}}h_{\texttt{dim}}=d$. 
 Then we have the hidden states of query $Q$, key $K$ and value $V$ as follows:
\begin{equation}
  { Q} = {\bf W}_q\bar{ E}_{\texttt{in}} , \quad { K} = {\bf W}_k\bar{ E}_{\texttt{in}} , \quad { V} = {\bf W}_v \bar{ E}_{\texttt{in}},  
\end{equation}
 and then we get the attention weights of parallel attention layers in $\mathcal{M}$, ${\bf W}_a$, and attention outputs $A$ as
\begin{equation}
    {\bf W}_a = \texttt{softmax}(QK^{\top}) \gamma \in \mathbb{R}^{h_{\texttt{attn}} \times h_{\texttt{attn}} \times l}, \quad A^{'} = {\bf W}_o( {\bf W}_a V)\in \mathbb{R}^{l \times h_{\texttt{attn}}h_{\texttt{dim}}},
\end{equation}
where $\gamma $ is the scaling factor. The attention outputs are then normalized again as $\bar{A}^{'} = \texttt{RMSNorm}(A^{'})$, then input to an MLP $\mathcal{M}_{\texttt{MLP}} = \{{\bf W}_{\texttt{gate}}, {\bf W}_{\texttt{up}}, {\bf W}_{\texttt{down}}\}$, such that we have the hidden state
\begin{equation}
    A = \texttt{RMSNorm}{\Large(} {\bf W}_{\texttt{down}}{\Large(} \sigma {\Large(} ({\bf W}_{\texttt{gate}}{\bar{A}^{'} })({\bf W}_{\texttt{up}}{\bar{A}^{'}  }){\Large)}{\Large)}{\Large)} \in \mathbb{R}^{ l \times h_{\texttt{attn}}h_{\texttt{dim}}},
\end{equation}
where $\sigma$ is the activation function.
Suppose there are $h$ parallel attention layers in $\mathcal{M}$, the attention outputs of the $i$th attention layer is $H_i$, so the final output of the transformer module $\mathcal{M}$ is 
\begin{equation}
    \bigoplus_{i=1}^h A_i = A_1 \oplus A_2 \oplus \cdots \oplus A_h \in \mathbb{R}^{h \times l \times h_{\texttt{attn}}h_{\texttt{dim}}}.
\end{equation}
Here, $\oplus$ denotes the direct sum, which means concatenation in the actual implementation. Since in the default setting, $h_{\texttt{attn}}h_{\texttt{dim}}=d$, we have the final hidden states output $\bigoplus_{i=1}^h A_i \in \mathbb{R}^{h \times l \times d} $.
\subsection{Additional Experiment Results}\label{sec:error}

The fitting details in~\cref{fig:eval} are clarified in~\cref{tab:fitting}.
\begin{table}[h]\small
\centering
\caption{Fitting perplexities in~\cref{fig:ppl} and accuracies in~\cref{fig:acc}.}\label{tab:fitting}
\begin{tabular}{m{4cm}|m{2.5cm}|c|c|c}
\toprule
\multirow{2}{*}{{\bf Perplexity} (\cref{fig:ppl})} & \multirow{2}{*}{{\bf Dataset}$(D)$}   & \multicolumn{3}{c}{$y=as+b \quad (y= \frac{\ln {\texttt{PPL}_0(D)}}{\ln{\texttt{PPL}(D)}}$}                            \\ \cline{3-5} 
                            &                            & {$a$} & {$b$} & RMSE \\ \midrule
Llama-3-8B-Instruct         & \multirow{2}{*}{WikiText2} & {-1.08}  & {0.96}  &   0.03   \\ \cline{1-1} \cline{3-5} 
Phi-3-mini-4k-Instruct      &                            & {-0.90}  & {1.02}  &    0.01  \\ \bottomrule

\end{tabular}

\vspace{15px}


\begin{tabular}{m{4cm}|m{2.5cm}|c|c|c}

\toprule
\multirow{2}{*}{{\bf Accuracy} (\cref{fig:acc})} & \multirow{2}{*}{{\bf Dataset}$(D)$}   & \multicolumn{3}{c}{$y=as+b\quad (y = \ln \frac{\texttt{acc}(D)}{\texttt{acc}_0(D)})$}                            \\ \cline{3-5} 
                            &                            & {$a$} & {$b$} & RMSE \\ \midrule
Llama-3-8B-Instruct         & \multirow{2}{*}{ARC-e} & {-2.14}  & {0.04}  &   0.05   \\ \cline{1-1} \cline{3-5} 
Phi-3-mini-4k-Instruct      &                            & {-1.84}  & {0.04}  &    0.04  \\ \midrule
Llama-3-8B-Instruct         & \multirow{2}{*}{ARC-c} & {-2.02}  & {-0.07}  &   0.09   \\ \cline{1-1} \cline{3-5} 
Phi-3-mini-4k-Instruct      &                            & {-1.88}  & {-0.01}  &    0.02  \\ \midrule
Llama-3-8B-Instruct         & \multirow{2}{*}{WinoGrande} & {-0.86}  & {-0.02}  &   0.02   \\ \cline{1-1} \cline{3-5} 
Phi-3-mini-4k-Instruct      &                            & {-0.66}  & {-0.02}  &    0.02  \\ \midrule
Llama-3-8B-Instruct         & \multirow{2}{*}{PIQA} & {-0.91}  & {-0.01}  &   0.03   \\ \cline{1-1} \cline{3-5} 
Phi-3-mini-4k-Instruct      &                            & {-0.90}  & {0.01}  &    0.01  \\ 
\bottomrule
\end{tabular}
\end{table}

\subsection{Limitations and Future Work}\label{app:future}
This paper gives explicit analytical relations between the pruned model performance and sparsity. However, these relations are derived from the direct representation dimension reduction, which is implemented by SliceGPT. The representation dimension may play similar roles in other pruning approaches (explicitly or implicitly), and it is worth investigating how it impacts their pruned model performance.

\end{document}